\documentclass{article}
\usepackage{ijcai23}

\usepackage{times}
\usepackage{url}
\usepackage[utf8]{inputenc}
\usepackage[small]{caption}
\usepackage{graphicx}
\usepackage{amsmath}
\usepackage{amsthm}
\usepackage{booktabs}
\usepackage{algorithm}
\usepackage{algorithmic}
\usepackage[switch]{lineno}
\usepackage{stfloats}

\usepackage{subfigure}
\usepackage[hidelinks]{hyperref}
\usepackage{nicefrac}       
\usepackage{microtype}      
\usepackage{xcolor}
\usepackage{multirow}
\usepackage{wrapfig}
\usepackage{array}
\usepackage{soul}
\usepackage{float}
\usepackage{bm}

\usepackage{wasysym}
\usepackage[ruled,vlined,algo2e]{algorithm2e}

\usepackage{color}
\usepackage{dsfont}	
\usepackage{footnote}
\usepackage{epstopdf}
 \usepackage{enumitem}
\usepackage{caption}
\usepackage{comment}

\usepackage{booktabs}       

\usepackage{epsfig}
\usepackage{amssymb}
\usepackage{adjustbox}
\usepackage[flushleft]{threeparttable}

\title{Part Aware Contrastive Learning for Self-Supervised Action Recognition}

\author{
Yilei Hua$^1$
\and
Wenhan Wu$^2$\and
Ce Zheng$^3$\and
Aidong Lu$^2$\and
Mengyuan Liu$^4$\and
Chen Chen$^3$\and
Shiqian Wu$^1$
\affiliations
$^1$School of Information Science and Engineering, Wuhan University of Science and Technology\\
$^2$University of North Carolina at Charlotte\\
$^3$Center for Research in Computer Vision, University of Central Florida\\
$^4$Peking University, Shenzhen Graduate School\\
\emails
\{hyl1997, shiqian.wu\}@wust.edu.cn,
\{wwu25, aidong.lu\}@uncc.edu,
cezheng@knights.ucf.edu,
chen.chen@crcv.ucf.edu,
nkliuyifang@gmail.com
}

\begin{document}

\maketitle

\begin{abstract}

In recent years, remarkable results have been achieved in self-supervised action recognition using skeleton sequences with contrastive learning. It has been observed that the semantic distinction of human action features is often represented by local body parts, such as legs or hands, which are advantageous for skeleton-based action recognition. This paper proposes an attention-based contrastive learning framework for skeleton representation learning, called SkeAttnCLR, which integrates local similarity and global features for skeleton-based action representations. To achieve this, a multi-head attention mask module is employed to learn the soft attention mask features from the skeletons, suppressing non-salient local features while accentuating local salient features, thereby bringing similar local features closer in the feature space. Additionally, ample contrastive pairs are generated by expanding contrastive pairs based on salient and non-salient features with global features, which guide the network to learn the semantic representations of the entire skeleton. Therefore, with the attention mask mechanism, SkeAttnCLR learns local features under different data augmentation views. The experiment results demonstrate that the inclusion of local feature similarity significantly enhances skeleton-based action representation. Our proposed SkeAttnCLR outperforms state-of-the-art methods on NTURGB+D, NTU120-RGB+D, and PKU-MMD datasets.The code and settings are available at this repository: \href{https://github.com/GitHubOfHyl97/SkeAttnCLR}{https://github.com/GitHubOfHyl97/SkeAttnCLR}
\end{abstract}

\begin{figure}[t]
  \centering
  \captionsetup{font=small}
  \includegraphics[width=1.0\linewidth]{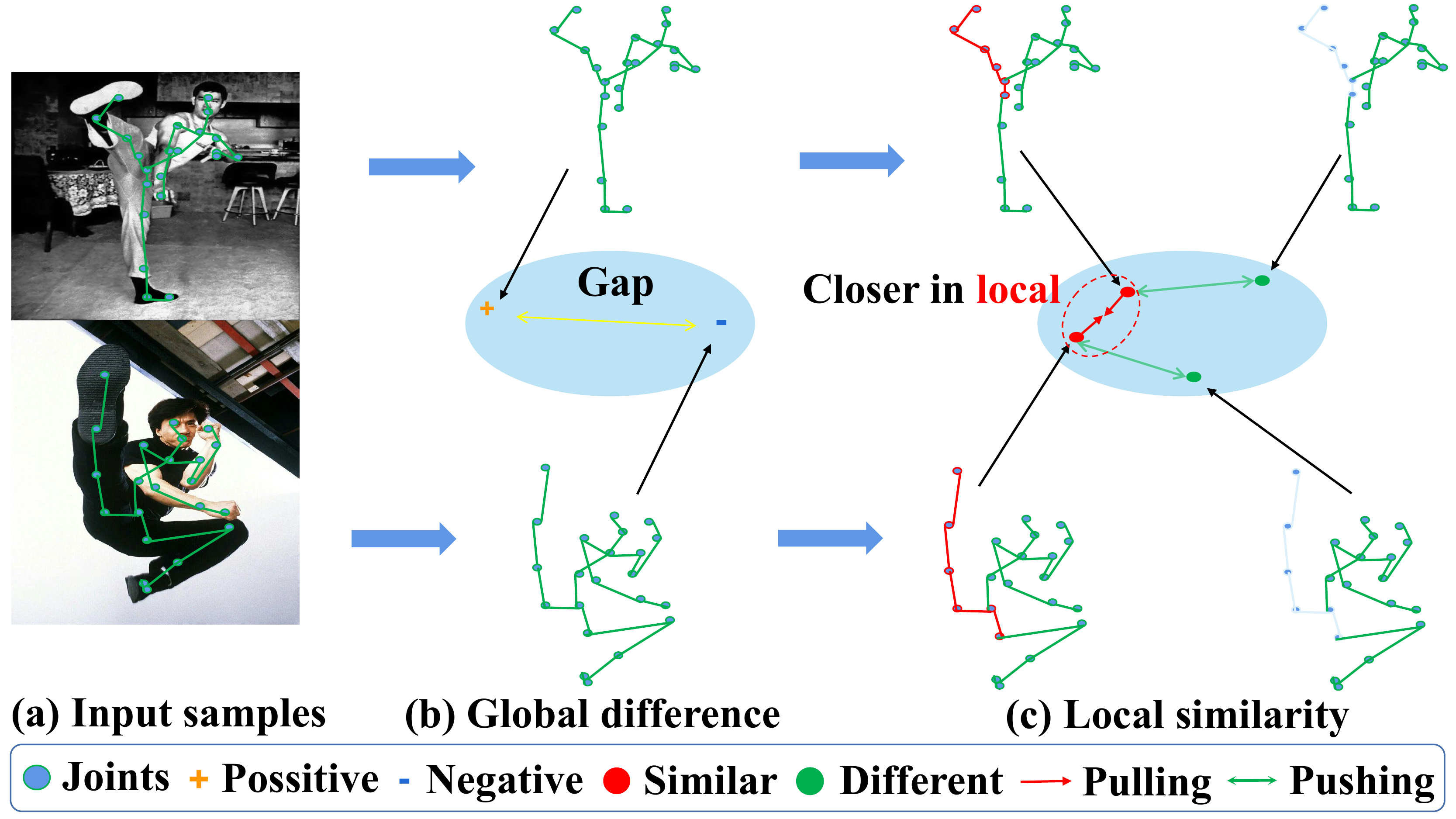}
   \caption{The motivation of our method. Although input skeletons (a) belong to the same movement category, there exists a significant gap between their skeleton sequences in feature space in (b). However, by considering local feature similarity, as shown by the red points in (c), the distance between two actions of the same semantic category becomes shorter in feature space. Therefore, we aim to extend local similarity-based contrastive learning to complement global contrastive learning, in order to bring samples with similar local features closer in feature space. Through the attention mechanism, the local similarity is discriminated and the network focuses on the changes in human action parts, such as leg movements (as highlighted in red). This approach is expected to be more conducive to learning local representations that are beneficial for accurate action recognition.}
   \label{fig1}
\end{figure}

\section{Introduction}


With the advancements in human pose estimation algorithms \cite{cao2017realtime,sun2019deep,zhengfeater,zheng2022lightweight}, skeleton-based human action recognition has emerged as an important field in computer vision. However, traditional supervised learning methods \cite{zhang2020semantics,ChiaraPlizzari2021SpatialTT} require extensive labeling of skeleton data, resulting in significant human effort. Thus, self-supervised learning has gained attention recently due to its ability to learn action representations from unlabeled data. Self-supervised learning has shown success in natural language \cite{kim2021self} and vision \cite{he2022masked}, leading researchers to explore self-supervised learning pre-training for human skeleton-based action recognition. 


The current skeleton-based contrastive learning framework has been developed from image-based methods, with researchers exploring cross-view learning \cite{li20213d} and data augmentation \cite{guo2022contrastive,zhang2022hierarchical}. In this work, we focus on instance discrimination via contrastive learning for self-supervised representation learning. Our proposed method emphasizes learning the representation of local features and generating more contrastive pairs from various local actions within a sample to improve contrastive learning performance. As depicted in Figure \ref{fig1}, such action samples are often challenging due to significant global representation differences, resulting in a considerable gap in the feature space. However, the distance between local features in the same action category is closer in the feature space, leading to semantic similarity. Local actions often determine semantic categories, making it desirable to consider local similarities in contrastive learning. In this study, we propose to improve previous works by addressing the following: 1) How to learn the relationship between local features and global features of human actions in skeleton-based self-supervised learning? 2) How to ensure that the contrastive learning network learns features with local semantic action categories?


In this paper, we propose SkeAttnCLR, a contrastive framework for self-supervised action recognition based on the attention mechanism. It addresses the issues of learning the relationship between local and global features of human actions and ensures that the contrastive learning network learns features with local semantic information. Motivated by Fig. \ref{fig1}, the proposed scheme consists of two parts: global contrastive learning and local contrastive learning. The global contrastive learning follows the spirit of SkeletonCLR \cite{li20213d}, which is used to learn the global structure information of the human skeleton. The local contrastive learning is developed to learn local action features with discriminative semantic information. Specifically, the attention mask module based on a multi-head self-attention mechanism \cite{vaswani2017attention} (MHSAM) is used to explore local features. This module divides skeleton action features into salient and non-salient areas at the feature level. Contrastive pairs are constructed for salient and non-salient features, as well as negative contrastive pairs between them to represent their oppositions in the contrastive learning model. This allows the network to learn key features with semantic distinction, regardless of whether they are embodied in salient or non-salient features.





The proposed method, SkeAttnCLR, presents a novel contrastive learning architecture that effectively learns the overall structure of human skeletal actions through global contrastive learning while also extracting key action features through local contrastive learning. As our SkeAttnCLR performs feature-level attention without interfering with the encoder structure, it can be applied to different encoder types, making it generalizable in extracting better action representations for downstream tasks. Our contributions are summarized as follows:

\begin{itemize}[leftmargin=*]
\item A novel contrastive learning architecture is presented, in which the overall structure of human skeletal actions are learned through global contrastive learning, and key action features  are extracted through local contrastive learning.

\item A global-local contrastive learning framework SkeAttnCLR that leverages the attention mechanism with local similarity for skeleton-based models is proposed.

\item We develop the Multi-Heads Attention Mask module to improve contrastive learning performance by generating ample contrastive pairs. This is achieved via salient and non-salient features.

\item The proposed method outperforms the state-of-the-art methods in most evaluation metrics and especially achieves an overall lead in comprehensive comparison with the baseline, which employs only global features.
\end{itemize}

\begin{figure*}[t]
  \centering
  \captionsetup{font=small}
  \includegraphics[width=0.85\linewidth]{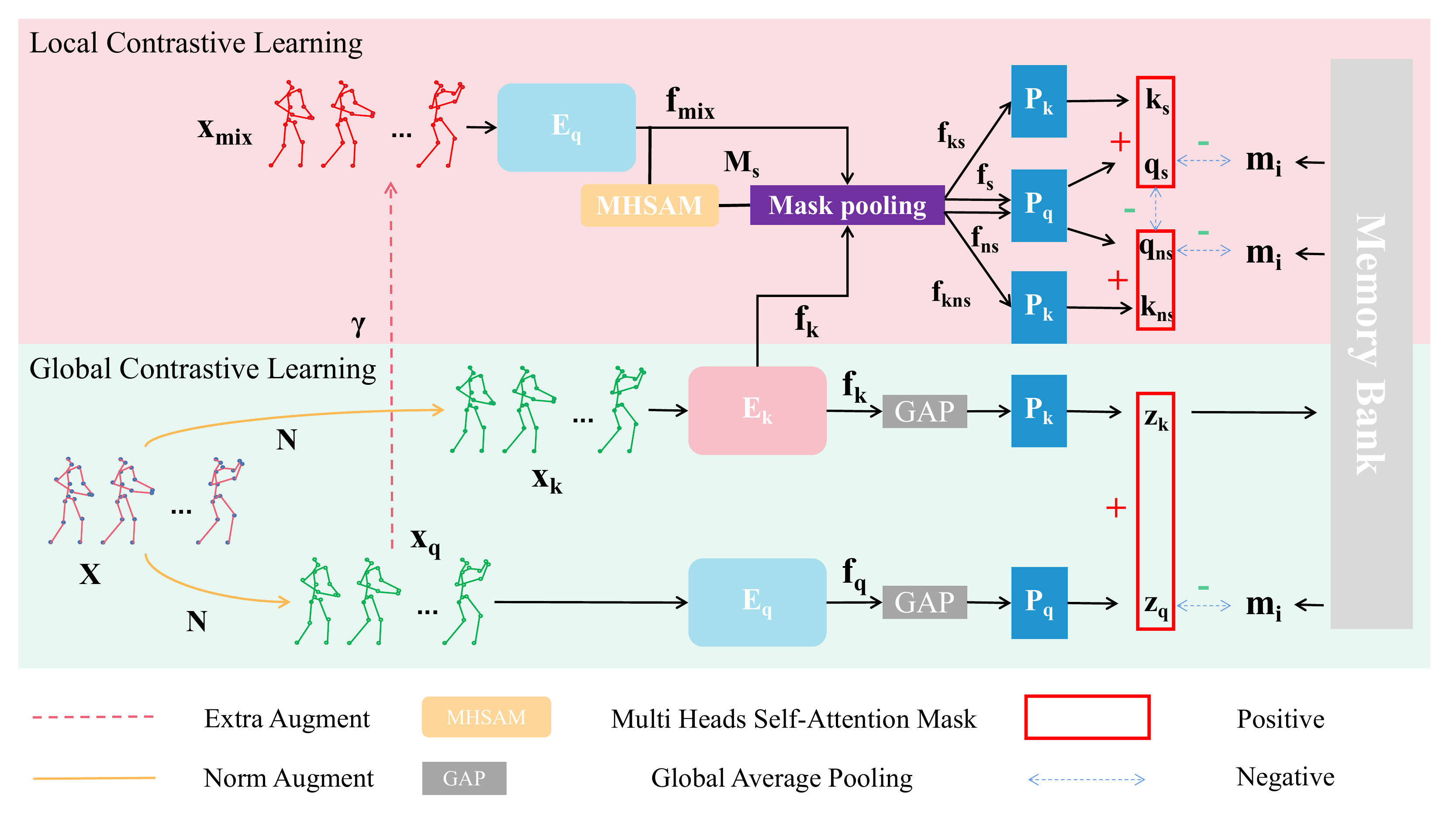}
    \caption{Architecture of the proposed SkeAttnCLR. First, in the global contrastive learning, original data $\mathrm{X}$ first obtains two different data augmented versions $x_{q}$ and $x_{k}$ after normal data augmentation $\mathrm{N}$ in Section \ref{subsec: Global Contrastive Learning}. Then, $x_{q}$ and $x_{k}$ input their respective encoders $\mathrm{E}_{q}$ and $\mathrm{E}_{q}$ to obtain feature embeddings $f_{q}$ and $f_{k}$ for the next step of calculation. Meanwhile, the $x_{q}$ is converted into $x_{mix}$ after extra data augmentation $\gamma$ in Section \ref{subsec: Local Contrastive Learning}. $x_{mix}$ gets its feature embedding $f_{mix}$ through $\mathrm{E}_{q}$. Finally, $f_{q}$ and $f_{k}$ are calculated for global contrastive learning as described in Section \ref{subsec: Global Contrastive Learning}. At the same time, $f_{mix}$ and $f_{k}$ obtain local feature embedding under the action of the soft mask $\mathrm{M}_{s}$ generated by the MHSAM module and perform local contrastive learning calculations according to Section \ref{subsec: Local Contrastive Learning}. As described in Section \ref{subsec: Global Contrastive Learning}, at each iteration, the $z_{k}$ of the current round is stored in the memory bank $M_{k}$. Then, according to the storage order, $m_{i}$ is generated to construct negative pairs.}
   \label{fig2}
\end{figure*}
\section{Related work}
\label{sec:related_work}

\subsection{Self-Supervised Representation Learning} 
Self-supervised learning is a method of learning data representation in a large amount of unlabeled data by setting a specific pretext task. There are many kinds of pretext tasks in self-supervised learning, which can be divided into two types: generative and discriminative. The examples, such as jigsaw puzzles \cite{wei2019iterative}, data restoration \cite{pathak2016context}, and Mask-based methods \cite{bao2021beit,he2022masked} that have been successful in image and language fields are generative self-supervised learning. Instance discrimination is a class of discriminative self-supervised learning pretext tasks. Instance discrimination based on contrastive learning \cite{he2020momentum} is a class of discriminative self-supervised learning pretext tasks. In recent years, the contrastive learning method of the Moco series \cite{he2020momentum,chen2020improved} stores the key vector by constructing a memory bank queue, and updates the encoder K through the momentum update mechanism. SimCLR \cite{chen2020simple} improves the performance of contrastive learning by adding additional MLP modules and embedding calculations with large batch sizes. In addition, BYOL \cite{richemond2020byol}, contrastive clustering \cite{li2021contrastive}, DINO \cite{caron2021emerging}, and SimSiam \cite{chen2021exploring} have also achieved promising results. In the recent work LEWEL \cite{huang2022learning}, the importance of local feature contrastive learning is emphasized for the first time in image tasks. The work of predecessors laid the foundation for our SkeAttnCLR, allowing us to go further on this basis.

\subsection{Skeleton-Based Action Recognition}  
Earlier human skeleton action recognition models based on deep learning are mainly designed based on RNN \cite{hochreiter2001gradient} and CNN \cite{ke2017new,li2017skeleton}. In recent years, due to the development of graph networks, human action skeleton recognition has begun to use GRU \cite{shi2017learning,su2020predict} or GCN-based \cite{li2019actional,liang2019three} models. At the same time, due to the recent success of the Transformer model in images and natural language, there have been many attempts to design a Transformer-based \cite{shi2020decoupled,ChiaraPlizzari2021SpatialTT} human skeleton action recognition model. ST-GCN \cite{yan2018spatial} is a widely used GCN-based human skeleton recognition model in recent years. It models the skeleton data structure from the perspective of Spatial-Temporal. In the experiment of this paper, we mainly use ST-GCN as the backbone encoder. In addition, in order to demonstrate the generalizability of our method, we also use the GRU-based BIGRU \cite{su2020predict} and Transformer-based DSTA \cite{shi2020decoupled} to conduct comparison experiments with the baseline.

\subsection{Contrastive Learning for Skeleton-Based Models Pre-training} 
 SkeleonCLR \cite{li20213d} is a simple contrastive learning framework designed on the basis of MocoV2 \cite{chen2020improved}. On this basis, CrossCLR \cite{li20213d} was proposed for multi-view contrastive learning to achieve cross-view consistency. AimCLR \cite{guo2022contrastive} and HiCLR \cite{zhang2022hierarchical} aim to expand more contrastive pairs under stronger data augmentation conditions to improve single-view contrastive learning performance. SkeleMixCLR \cite{chen2022contrastive} relies on the unique skeleton mixing data augmentation to design a targeted contrastive learning framework. It is noted that there is a lack of consideration of how to use the local features of human motion in the existing skeleton-based contrastive learning methods. The data augmentation method in the SkeleMixCLR locally mixes real human action parts to explore local feature combinations. However, this method needs to be marked at the feature level according to the Spatial-Temporal position of the data mixture, which requires the data to maintain Spatial-Temporal consistency after downsampling. Hence SkeleMixCLR is not conducive to extending to other backbones. It is desirable to propose a simple and generalizable local contrastive learning method.

\section{SkeAttnCLR}
\label{sec:SkeAttnCLR}
As aforementioned, the local information of human motion has not been fully mined and emphasized. In this study, We attempt to extend local contrastive learning based on feature-level local similarity to global contrastive learning. We also use the attention mask generated by MHSAM to divide our defined attention salient features and non-salient features at the feature level. Then, contrastive pairs are constructed using the relations within local features, and between local and global features for the pretext task of instance discrimination.

SkeAttnCLR is shown in Fig. \ref{fig2}, which is a method built on a single view. In the global part, we follow the basic design of SkeletonCLR \cite{li2021contrastive}. The input to the local contrastive learning part comes from further data augmentation of the global part query input. In the local contrastive learning part, we divide the feature vectors obtained from encoder $\mathrm{E}_{q}$ or encoder $\mathrm{E}_{k}$ into attention-salient and non-salient embeddings through the MHSAM module. Finally, local-to-local, global-to-global, and local-to-global contrastive pairs are constructed between the global and local embeddings.

\subsection{Global Contrastive Learning}
\label{subsec: Global Contrastive Learning}
In this section, we introduce the specific details of global contrastive learning, laying the foundation for the subsequent introduction of local contrastive learning. 

\textbf{Data Augmentation.} 
For the input data of global contrastive learning, we adopt Shear and Crop\cite{li20213d} as the augmentation strategy. We refer to this part of the data augmentation combination as normal data augmentation $\mathrm{N}$. N randomly converts the read skeleton sequence $\mathrm{X}$ into two different data-augmented versions $x_{q}$ and $x_{k}$ as positive pairs.

\textbf{Global Contrastive Learning Module.} 
As shown in Fig. \ref{fig2}, the two encoders $\mathrm{E}_{q}$ and $\mathrm{E}_{k}$ respectively embed $x_{q}$ and $x_{k}$ into the feature space:  $f_{q}=\mathrm{E}_{q}\left(x_{q}; \theta_{q}\right)$ and $f_{k}=\mathrm{E}_{k}\left(x_{k}; \theta_{k}\right)$, where $f_{q}, f_{k} \in \mathbb{R} ^ {n \times C_f}$. Among them, $E_{k}$ follows $E_{q}$ to update the parameters $\theta_{k}$ through the momentum update mechanism: $\theta_{k} \leftarrow M \theta_{k}+(1-M) \theta_{q}$, where $\mathrm{M}$ is a momentum coefficient. 
We apply a dynamic momentum coefficient that changes according to the number of global iterations following BYOL \cite{richemond2020byol}, the formula is as follows: 
$M=1-\left(1-M_{0}\right) \cdot(\cos{(\pi \cdot {iter/{iter_{max}}}})+1)/2$.

Then, after $f_{q}$ and $f_{k}$ are processed by global average pooling, they are respectively input into predictor $\mathrm{P}_{q}$ and $\mathrm{P}_{k}$ to obtain the output embedding $z_{q}$ and $z_{k}$: $z_{q}=\mathrm{P}_{q}\left(f_{q}\right)$ and $z_{k}=\mathrm{P}_{k}\left(f_{k} \right)$, where $z_{q}, z_{k} \in \mathbb{R} ^ {C_z}$. $p_{k}$ is the momentum updated version of $p_{q}$. 

To construct negative pairs, a queue $M_{k}=[m_{i}]^{K}_{i=1}$ named memory bank is used to store previous embeddings $z_{k}$. At each iteration, $M_{k}$ provides a large number of negative pairs for contrastive learning. In each iteration, $z_{k}$ is calculated by the previous samples stored in $M_{k}$ which is dequeued according to the storage order to obtain $m_{i}$, forming a large number of negative pairs with the newly calculated $z_{q}$.

In the global contrastive learning part, we adopt InfoNCE \cite{oord2018representation} for global instance discrimination of skeleton actions:
 \begin{equation}
L_{\text {Info }}=-\log \left(\frac{\exp (z_{q} \cdot z_{k} / \tau)}{\exp (z_{q} \cdot z_{k} / \tau)+Neg(z_{q})}\right)
\end{equation} 
$z_{q}$, $z_{k}$ and $m_{i}$ are all normalized. $\tau$ is a hyperparameter (set to 0.2 in experiments). $Neg(z)=\sum_{i=1}^{K} \exp \left(z \cdot m_{i} / \tau\right)$ denotes the similarity between embedding $z$ and $m_{i}$ from memory bank. In this loss function, the global distance between samples is calculated by the dot product. 

\begin{figure}[t]
  \centering
  \captionsetup{font=small}
  \includegraphics[width=0.99\linewidth]{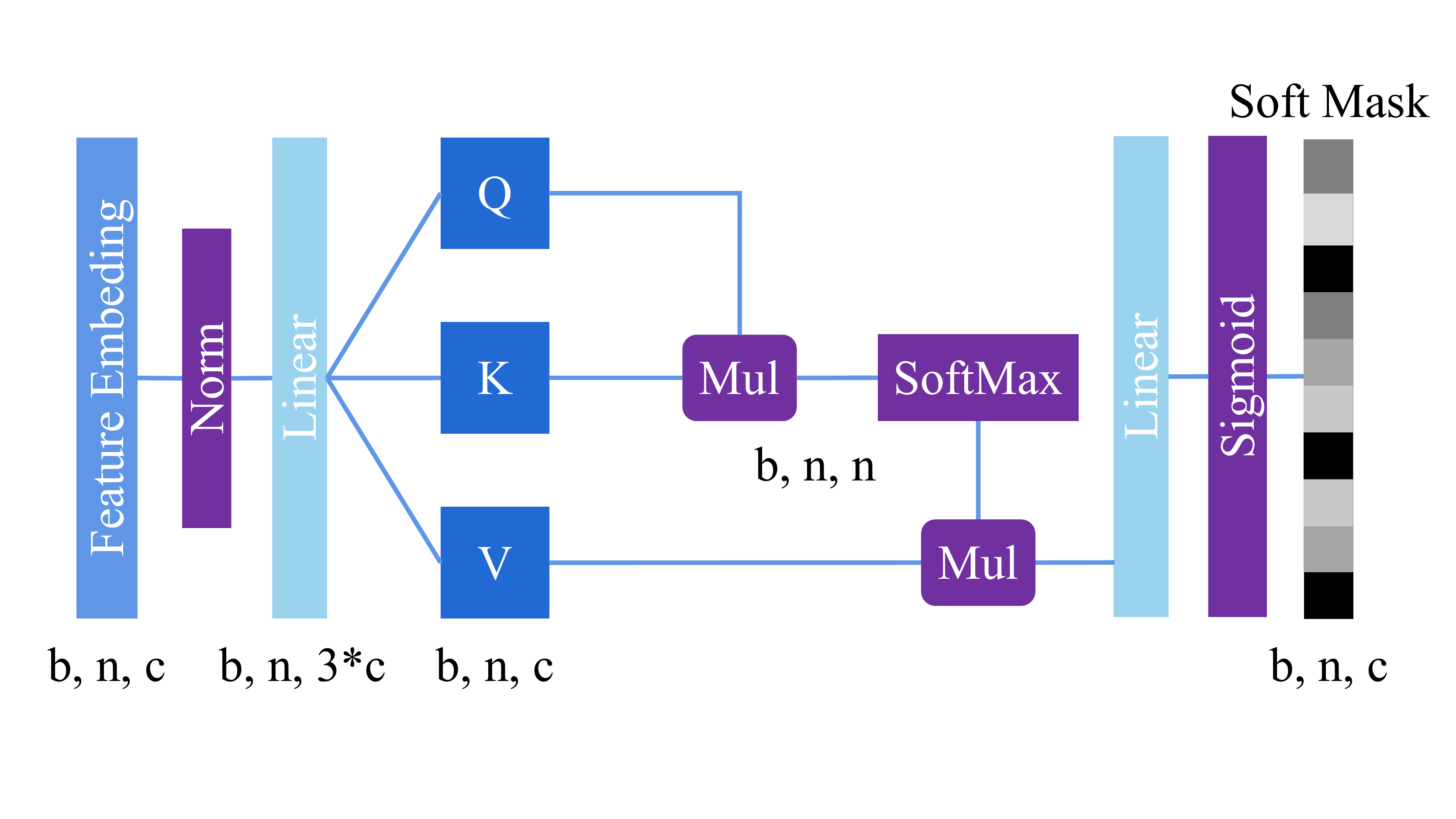}
   \caption{The Multi Head Self-attention Mask Module (MHSAM),  whose input is feature-level embedding, and the output is a soft mask, which is also regarded as an attention-weight matrix.}
   \label{fig3}
\end{figure}

\subsection{Local Contrastive Learning}
\label{subsec: Local Contrastive Learning}
Local contrastive learning builds upon the concept of global contrastive learning by specifically focusing on mastering the structural features of human action sequences.
Extra data augmentation can help the model explore more local feature combinations. The samples generated by the extra data augmentation get new feature embeddings through $\mathrm{E}_{q}$, and we use these feature embeddings to construct positive pairs with $z_{k}$, which locally brings similar samples closer. Conversely, the negative pairs of embeddings obtained from local features and $m_{i}$ stored in the memory bank will pull away samples that lack local similarities. In addition, we construct an opposite relationship between attention-salient features and non-attention-salient features through two methods of masking and setting up negative pairs. It not only expands the number of negative pairs in contrastive learning but also allows the network to learn to focus on parts with key action semantics.

In this section, we first perform extra data augmentation on the basis of $x_{q}$, then we use MHSAM to divide the feature embeddings of the local contrastive learning, and finally, we construct rich contrastive pairs between the divided local embeddings in pursuit of better local feature exploration.

\textbf{Extra Augmentation.} 
According to AimCLR \cite{guo2022contrastive}, stronger data augmentation is beneficial for learning human action features under certain methods. Therefore, in the part of local contrastive learning, we apply an extra data augmentation $\gamma$ on the basis of normal data augmentation results to explore the possibility of more human action features. Since data mixing augmentation \cite{chen2022contrastive} can randomly combine parts of different human motion samples, it is more conducive to exploring local features than other noise addition and filtering methods. Therefore, we apply data mixing data augmentation here as our extra augmentation. In the part of local comparison learning, $x_{q}$ is randomly transformed into $x_{mix}$ after data mixing augmentation through $\gamma$.

\textbf{Multi Head Self Attention Mask.} 
With the success of transformer \cite{vaswani2017attention} in various tasks, the multi-head self-attention mechanism has also attracted much attention as a part of the transformer. Due to its unique query-matching mechanism, the multi-head self-attention mechanism is able to estimate the correlation between features from multiple angles, and mine the connection between local features of actions at the feature level. The MHSAM module is shown in Fig. \ref{fig3}. The input to this module is a tensor of size $b \times n \times c$, where $b$ represents the batch size, $n$ is the length of a single data, and $c$ is the number of data channels.

MHSAM is designed to embed the encoder output at the feature level. Since the Encoder in the local comparative learning shares parameters with the $\mathrm{E}_{q}$ of the global comparative learning, so we have $f_{mix}=\mathrm{E}_{q}\left(x_{mix}; \theta_{q}\right)$, where $f_{mix} \in \mathbb{R} ^ {n \times C_f}$. The $\mathrm{~Q}$, $\mathrm{~K}$ and $\mathrm{~V}$ of the multi-head self-attention mechanism are calculated by the following formula:
 \begin{equation}
\mathrm{~Q}, \mathrm{~K}, \mathrm{~V}=\operatorname{Linear}\left(f_{mix}\right)=f_{mix}w_{q}, f_{mix}w_{k}, f_{mix}w_{v}
\end{equation} 
$w_{q}$, $w_{k}$ and $w_{v}$ is the parameter matrix of the linear network. Then attention feature is described by the softmax function:
 \begin{equation}
x_{attn}=\operatorname{softmax}\left(\frac{Q K^{T}}{\sqrt{d_{k}}}\right) V
\end{equation} 
where $\frac{1}{\sqrt{d_{k}}}$ is the normalization scaling factor. After calculation, $x_{attn}$ is put into a simple linear network projection for adjustment. Finally, there is a Sigmoid function to generate a soft mask $\mathrm{~M}_{s}$. The formula is as follows:
\begin{equation}
M_{s}=\operatorname{Sigmoid}(\lambda \cdot \operatorname{proj}(x_{attn}))
\end{equation} 
where $\lambda$ is a hyperparameter that adjusts the tolerance of neutral features. The larger the value of $\lambda$, the easier the value of the mask tends to be polarized.

\begin{figure}[t]
  \centering
  \captionsetup{font=small}
  \includegraphics[width=1.0\linewidth]{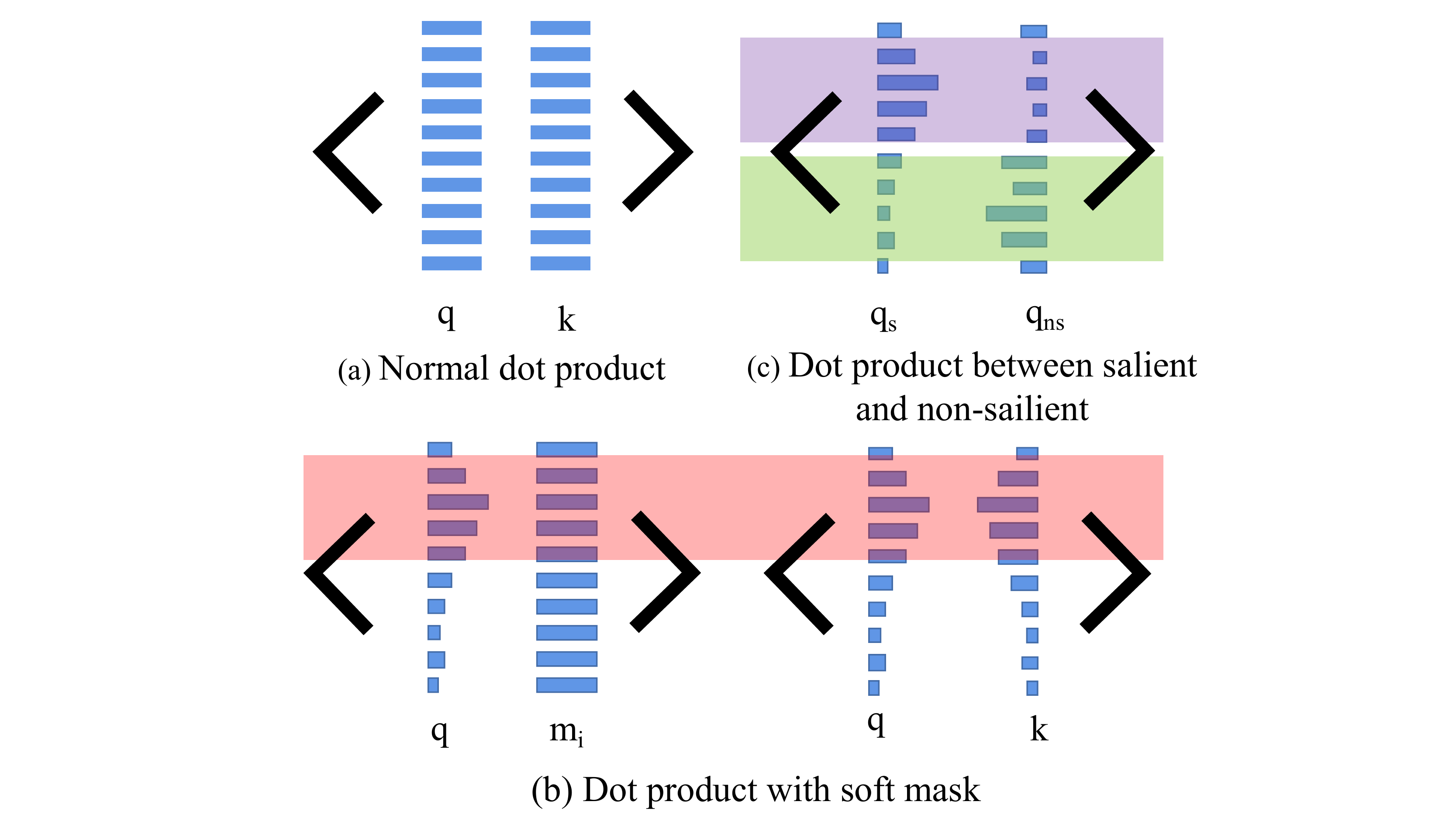}
   \caption { (a) The normal dot product in the calculation of $L_{\text {Info }}$. (b) and 
    (c) show the attention affected dot product in the calculation of $L_{s}$ and $L_{ns}$.}
   \label{fig4}
\end{figure}

\textbf{Local Contrastive Learning Module.} 
After obtaining the mask, we use it to separate the feature-level embedding extracted by the encoder into salient and non-salient features. Following the mask pooling procedure, inputs $f_{mix}$ and $f_{k}$ are transformed in both salient and non-salient ways to get $f_{s}, f_{ns}, f_{ks}$ and $f_{kns}$.  The formula of mask pooling is as follows:
\begin{equation}
M_{n s}=I-M_{s} 
\end{equation} 
\begin{equation}
f_{s}=\left( \sum^{n} f_{mix} \cdot M_{s}\right) / n, 
f_{ks}=\left( \sum^{n} f_{k} \cdot M_{s}\right) / n
\end{equation} 
\begin{equation}
f_{ns}=\left( \sum^{n} f_{mix} \cdot M_{ns}\right) / n ,
f_{kns}=\left( \sum^{n} f_{k} \cdot M_{ns}\right) / n
\end{equation} 
where $I$ is the identity matrix, and $n$ is the data length of the dimension that needs to be pooled. 

After getting the embeddings from mask pooling, we have $q_{s}=\mathrm{P}_{q}\left(f_{s}\right)$, $q_{ns}=\mathrm{P}_{q}\left(f_{ns}\right)$, $k_{s}=\mathrm{P}_{k}\left(f_{ks}\right)$ and $k_{ns}=\mathrm{P}_{k}\left(f_{kns}\right)$. In local contrastive learning, we make $q_{s}$ and $k_{s}$ a positive pair, $q_{ns}$ and $k_{ns}$ a positive pair, and set $q_{s}$ and $q_{ns}$ a negative pair. Then, the loss functions of salient features and non-salient features are extended by Section \ref{subsec: Global Contrastive Learning} as:
\begin{scriptsize}
\begin{equation}
L_{s}=-\log \left(\frac{\exp \left(q_{s} \cdot k_{s} / \tau\right)}{\exp \left(q_{s} \cdot k_{s} / \tau\right)+\exp \left(q_{s} \cdot q_{ns} / \tau\right)+Neg(q_{s})}\right) 
\end{equation}
\end{scriptsize}

\begin{scriptsize}
\begin{equation}
L_{ns}=-\log \left(\frac{\exp \left(q_{n s} \cdot k_{n s} / \tau\right)}{\exp \left(q_{n s} \cdot k_{n s} / \tau\right)+\exp \left(q_{n s} \cdot q_{s} / \tau\right)+Neg(q_{ns})}\right)
\end{equation} 
\end{scriptsize}

In the estimation of $L_{s}$ and $L_{ns}$, We utilize attention to leverage the calculation of local similarity as shown in Fig. \ref{fig4}. 
We assume that the general dot product is shown in Fig. \ref{fig4} (a).
The mask generated by MHSAM highlights the parts that are concerned at the feature level so that the result of the dot product operation tends to highlight the local similarity in Fig. \ref{fig4} (b).
The dot product between $q_{s}$ and $q_{ns}$ is shown in Fig. \ref{fig4} (c), which represents the opposition between them.
Besides, We still use the memory bank $M_{k}$ that stores global feature embeddings to provide negative pairs in local contrastive learning, thus it is worth noting that the dot product of local-to-global also emphasizes the local similarity as local-to-local does, as shown in Fig. \ref{fig4} (b).
Finally, the total loss function for local contrastive learning can be expressed as:
\begin{equation}
L_{\text {local }}=\mu L_{s}+(1-\mu) L_{ns}
\end{equation} 
where $\mu \in(0,1)$.

\subsection{The Overall Objective of SkeAttenCLR}
\label{subsec: The Overall Objective of SkeAttenCLR}
SkeAttnCLR estimates the distance of global features in the feature space between samples through global contrastive learning for global instance discrimination and also computes the distance of local features through local contrastive learning for instance discrimination. Combining local and global contrastive learning, SkeAttnCLR can be optimized by the following loss function:

\begin{equation}
L=L_{\text {global }}+L_{\text {local }}=L_{\text {info }}+L_{\text {local }}
\end{equation} 

\section{Experiments}
\label{sec:experiments}
\subsection{Datasets}
\label{sec:datasets}
We select the most widely used NTU series dataset for experimental evaluation. The NTU dataset contains a wide range of human action categories and has a unified and standardized data processing code in use for many years, which ensures fairness when compared with previous methods.

\textbf{NTU-RGB+D 60 (NTU-60).}
NTU-60  \cite{shahroudy2016ntu} is a large-scale skeleton dataset for human skeleton-based action recognition, containing 56,578 videos with 60 actions and 25 joints for each human body. The dataset includes two evaluation protocols: the Cross-Subject (X-Sub) protocol, which divides data by subject with half used for training and half for testing, and the Cross-View (X-View) protocol, which uses different camera views for training. The testing samples are captured by cameras 2 and 3 for training, and samples from camera 1 are used for testing. 

\textbf{NTU-RGB+D 120 (NTU-120).}
NTU-120  \cite{liu2019ntu} is an expansion dataset of NTU-60, containing 113,945 sequences with 120 action labels.  It also offers two evaluation protocols, the Cross-Subject (X-Sub) and Cross-Set (X-Set) protocols. In X-Sub, 53 subjects are used for training and 53 subjects are used for testing, while in X-Set, half of the setups are used for training (even setup IDs) and the remaining setups (odd setup IDs) are used for testing.

\textbf{PKU Multi-Modality Dataset (PKU-MMD).}
PKU-MMD \cite{liu2020benchmark} is a substantial dataset that encompasses a multi-modal 3D comprehension of human actions, containing around 20,000 instances and 51 distinct action labels. It is split into two subsets for varying levels of complexity: Part I is designed as a simpler version, while Part II offers a more challenging set of data due to significant view variation. 

\subsection{ Experimental Settings}
\label{subsec: Experimental settings}
Our experiments mainly use the SGD optimizer \cite{ruder2016overview} to optimize the model. For all contrastive learning training, we use a learning rate of 0.1, a momentum of 0.9, and a weight decay of 0.0001 for a total of 300 epochs for training, and adjust the basic learning rate to one-tenth of the original at the 250th epoch. In addition, our data processing employs human skeleton action sequences with a length of 64 frames, and the batch size is 128. We choose ST-GCN \cite{yan2018spatial} as the main backbone of our experiments for a fair comparison, as it is the most widely adopted method in existing skeleton-based action recognition approaches.
Meanwhile, We also provide the detailed settings for all backbones in the experiments and t-Distributed Stochastic Neighbor Embedding (t-SNE) \cite{van2008visualizing} visualization results in the \ref{sec: detailed architecture} and \ref{sec: visualization}. 

\textbf{KNN evaluation protocol. }
During the contrastive learning training process, we directly use a KNN cluster every 10 epochs to cluster the feature embeddings extracted by the encoder and evaluate the clustering accuracy on the test set. Finally, the model with the highest KNN result is selected to participate in other experiments.

\textbf{Linear evaluation protocol. }
Linear evaluation is the most commonly used evaluation method for downstream classification tasks. Its usual practice is to freeze the parameters of the backbone encoder trained by self-supervised learning, and use supervised learning to train a linear fully connected layer classifier in the test. In the experiment, we use the SGD optimizer with a learning rate of 3 to train for 100 epochs and adjust the base learning rate at 60th epoch.

\textbf{Finetune evaluation protocol. }
Different from the linear evaluation, finetune evaluation does not freeze the parameter update of the encoder. We use SGD with an initial learning rate of 0.05 for optimization. In this experiment, the dynamic learning rate adjustment that comes with PyTorch \cite{paszke2019pytorch} is applied, which automatically adjusts when the loss does not converge.

\textbf{Semi-finetune evaluation protocol. }
The difference between semi-finetune  and finetune evaluation is that the former only uses a few labeled data for training. 
We experimented with 1$\%$ labeled data and 10$\%$ labeled data respectively, and the optimizer settings follow finetune evaluation.

\begin{table*}[htp]
\centering
\small
{

\begin{tabular}{c|c|cc|cc|cc}
\hline
        &       & \multicolumn{2}{c|}{NTU-60} & \multicolumn{2}{c|}{NTU-120}   & \multicolumn{2}{c}{PKU-MMD} \\  \cline{3-8}
 \multirow{-2}{*}{Method}        &   \multirow{-2}{*}{Stream}       &\multicolumn{1}{l}{Xsub}                  & \multicolumn{1}{l|}{Xview}                 & \multicolumn{1}{l}{Xsub}                &  \multicolumn{1}{l|}{Xset}
         & \multicolumn{1}{l}{Part I}                &  \multicolumn{1}{l}{Part II}         \\ \hline
Baseline      & J  & 68.3                                          & 76.4                                         & -                      & -      & -         & -       \\
Baseline*     & J  & 72.0 \color{red}{+3.7}                                  & 79.0 \color{red}{+2.6}                                & 51.0                  & 61.7        & 80.3  & 39.1           \\
\textbf{Ours}          & J  &  \textbf{80.3} \color{red}{+12.0} & \textbf{86.1}\color{red}{+9.7} & \textbf{66.3}\color{red}{+15.1}  & \textbf{74.5}\color{red}{+12.8} & \textbf{87.3}\color{red}{+7.0} & \textbf{52.9}\color{red}{+23.8}\\ \hline
Baseline      & M & 53.3                                          & 50.8                                         & -                      & -  & -  & -                   \\
Baseline*     & M & 56.5\color{red}{+3.2}                                  & 57.2\color{red}{+6.4}                                 & 46.1                  & 43.8       & 66.5  & 14.0           \\
\textbf{Ours}           & M & \textbf{63.9}\color{red}{+10.6}                        & \textbf{58.7}\color{red}{+7.9}                        & \textbf{49.9}\color{red}{+3.8} & \textbf{59.3}\color{red}{+15.5}  & \textbf{72.2}\color{red}{+5.7} & \textbf{32.7}\color{red}{+18.7}\\ \hline
Baseline     & B   & 69.4                                          & 67.4                                         & -                      & -       & -  & -               \\
Baseline*     & B   & 66.0\color{cyan}{-3.4}                                 & 69.0\color{red}{+1.6}                                 & 51.1                  & 56.3      & 79.7  & 21.5           \\
\textbf{Ours}           & B   & \textbf{76.2}\color{red}{+6.8}                        & \textbf{76.0}\color{red}{+8.6}                        & \textbf{63.0}\color{red}{+11.9} & \textbf{67.3}\color{red}{+11.0}    & \textbf{87.3}\color{red}{+7.6}   & \textbf{37.7}\color{red}{+16.2}  \\ \hline
Baseline     & 3S     & 75.0                                             & 79.8                                           & -                      & -        & -   & -               \\
Baseline*    & 3S     & 75.9 \color{red}{+0.9}                                           & 79.8 \color{green}{0}                                            & 65.0                      & 65.9        & 85.3       & 38.8       \\
\textbf{Ours}           & 3S     & \textbf{82.0}\color{red}{+6.1}                                             & \textbf{86.5}\color{red}{+6.7}                                            & \textbf{77.1}\color{red}{+12.1}                        & \textbf{80.0}\color{red}{+14.1}    & \textbf{89.5}\color{red}{+4.2}    & \textbf{55.5}\color{red}{+16.7}   \\ \hline
\end{tabular}
}
\caption{Linear evaluation comparisons with the baseline using the same backbone, where J, M, and B indicate joint, motion, and bone,  3S means three streams fusion, $*$ indicates that results obtained with our settings.}
\label{table1}
\end{table*}

\subsection{Result Comparison}
To validate the effectiveness of our method, we compare it with other methods of linear evaluation, KNN evaluation, finetune evaluation, and semi-finetune evaluation protocols. To facilitate fair comparisons, we mainly choose similar methods that use ST-GCN (backbone network) and achieve SOTA in recent years for comparison.

\textbf{Comparisons with baseline.}
Our method is compared with the baseline(SkeletonCLR\cite{li20213d}) on NTU-60, NTU-120 and PKU-MMD datasets when ST-GCN is used as the backbone encoder. The results are shown in Table \ref{table1}. In order to demonstrate the generalizability of our method, we additionally use BIGRU \cite{su2020predict} and transformer (DSTA) \cite{shi2020decoupled} as the backbone encoder on the NTU-60 dataset for comparison with the baseline. The results are shown in Table \ref{table2}. As we can see from Table \ref{table1} and Table \ref{table2}, our method has a comprehensive improvement compared to the baseline on different datasets and different backbone encoders. The experiments demonstrate the effectiveness of our method adding local contrastive learning on a global basis. 
\begin{table}[htp]
\centering
\small
{

\begin{tabular}{c|c|c|c|c}
\hline
              &       & ST-GCN                                                                                      & BIGRU                                                     & Transformer                                               \\ \cline{3-5} 
\multirow{-2}{*}{Method} & \multirow{-2}{*}{Stream} &  Xsub               & Xsub &Xsub  \\ \hline
Baseline   &                          & 68.3                                                          & -                                               & -                                                 \\
Baseline*   &                          & 72.0\color{red}{+3.7}                                                                   & 64.8                                            & 54.5                                            \\
\textbf{Ours}         & \multirow{-3}{*}{J}  & \textbf{80.3}\color{red}{+12.0}  & \textbf{72.7}\color{red}{+7.9}    & \textbf{71.3}\color{red}{+16.8}           \\ \hline
Baseline   &                          & 53.3                                                                                  &-                                                    & -                                                   \\
Baseline*   &                          & 56.5\color{red}{+3.2}                                                                  & 58.4                                            & 45.0                                           \\
\textbf{Ours}        & \multirow{-3}{*}{M} & \textbf{63.9}\color{red}{+10.6}                             & \textbf{74.9}\color{red}{+16.5}        & \textbf{52.3}\color{red}{+7.3}         \\ \hline
Baseline   &                          & 69.4                                                                                 & -                                                    & -                                                    \\
Baseline*  &                          & 66.0 \color{cyan}{-3.4}                                                                  & 63.0                                            & 51.0                                          \\
\textbf{Ours}         & \multirow{-3}{*}{B}   & \textbf{76.2}  \color{red}{+6.8}                                              & \textbf{69.2} \color{red}{+6.2}            & \textbf{73.7} \color{red}{+22.7}                                                    \\ \hline
\end{tabular}
}
\caption{Linear evaluation comparisons with different backbones on NTU-60 dataset. J, M, and B indicate joint, motion, and bone.}
\label{table2}
\end{table}

\begin{table}[htp]
\centering
\small
{ 
\begin{tabular}{c|c|cc|cc}
\hline
        &  & \multicolumn{2}{c|}{NTU-60} & \multicolumn{2}{c}{NTU-120}    \\  \cline{3-6}
 \multirow{-2}{*}{Method}        &   \multirow{-2}{*}{Stream}         & \multicolumn{1}{l}{Xsub}                  & \multicolumn{1}{l|}{Xview}                 & \multicolumn{1}{l}{Xsub}                &  \multicolumn{1}{l}{Xset} \\ \hline
SkeletonCLR        & J                     & 68.3                         & 76.4                         & 56.8                        & 55.9                         \\
CrossCLR          & J                  & 72.9                         & 79.9                         & -                           & -                            \\
AimCLR             & J                   & 74.3                         & 79.7                         & 63.4                        & 63.4                         \\
HiCLR             & J              & 77.6                         & 82.4                         & -                           & -                            \\
SkeleMixCLR     & J                   & 79.6                         & 84.4                         & {\color[HTML]{FF0000} 67.4} & 69.6                         \\
\textbf{Ours}      & J                   & {\color[HTML]{FF0000} 80.3} & {\color[HTML]{FF0000} 86.1} & 66.3                        & {\color[HTML]{FF0000} 74.5} \\ \hline
SkeletonCLR          & 3S          & 77.8                         & 83.4                         & 67.9                        & 66.7                         \\
AimCLR            & 3S       & 78.9                         & 83.8                         & 68.2                        & 68.8                         \\
HiCLR           & 3S  & 80.4                         & 85.5                         & 70                          & 70.4                         \\
SkeleMixCLR  & 3S       & 81                           & 85.6                         & 69.1                        & 69.9                         \\
\textbf{Ours}    & 3S       & {\color[HTML]{FF0000} 82.0} & {\color[HTML]{FF0000} 86.5} & {\color[HTML]{FF0000} 77.1} & {\color[HTML]{FF0000} 80.0} \\ \hline
\end{tabular}
}
\caption{Linear evaluation comparisons with other methods using the same backbone, J indicates joint, 3S means three streams fusion.}
\label{table3}
\end{table}
\textbf{Comparisons with previous works.}
For a fair comparison, we mainly select works that also mainly use ST-GCN for experiments in recent years and achieve SOTA to compare with our method. The experimental results are shown in Table \ref{table3}, our method is in an advantageous position in most comparisons. Especially in the comparison of three-stream results under the NTU-120 dataset, we have achieved a comparative advantage of more than 7$\%$. In the next analysis of KNN results, we combine the results of the linear evaluation to analyze the reason why our xsub single-stream in NTU-120 does not reach the best. Notably, we have a considerable improvement in the NTU-120 dataset with three streams of data ensemble, which indicates that SkeAttnCLR performs better with the fusion of joint-bone-motion streams.

\textbf{KNN evaluation results.}
As shown in Table \ref{table4}, in the KNN evaluation comparison with similar methods, our method achieves SOTA in most indicators. Based on the results of the linear evaluation, we speculate that SkeAttnCLR performs worse in xsub due to the increasing number of categories in the NTU-120 dataset, and the skeleton captured by xsub is not as good as that of xset, which leads to bias in local similarity for each category.

\begin{table}[htp]

\centering
\small
{ 
\begin{tabular}{c|cc|cc}
\hline
        & \multicolumn{2}{c|}{NTU-60} & \multicolumn{2}{c}{NTU-120}    \\  \cline{2-5}
 \multirow{-2}{*}{Method}        &   \multicolumn{1}{l}{Xsub}                  & \multicolumn{1}{l|}{Xview}                 & \multicolumn{1}{l}{Xsub}                &  \multicolumn{1}{l}{Xset}
      \\ \hline
SkeletonCLR  & 60.7                        & 64.8                        & 42.9                        & 41.9                         \\
AimCLR  & 63.7                        & 71.0                          & 47.3                        & 48.9                         \\
HiCLR  & 67.3                        & 75.3                        & -                           & -                            \\
SkeleMixCLR  & 65.5                        & 72.3                        & {\color[HTML]{FF0000} 48.3} & 49.3                         \\
\textbf{Ours}   & {\color[HTML]{FF0000} 69.4} & {\color[HTML]{FF0000} 76.8} & 46.7                       & {\color[HTML]{FF0000} 58.0} \\ \hline
\end{tabular}
}
\caption{KNN evaluation results on NTU-RGB+D dataset.}
\label{table4}
\end{table}


\textbf{Semi-finetune evaluation results.}
The experimental results of semi-finetune are shown in Table \ref{table5}, which show that the superiority of our method is not limited by the amount of labeled data.
\begin{table}[htp]
\centering
\small
{ 
\begin{tabular}{c|c|cc}
\hline
     &   & \multicolumn{2}{c}{NTU-60} \\  \cline{3-4}
 \multirow{-2}{*}{Method}        & 
  \multirow{-2}{*}{Label}        &\multicolumn{1}{l}{Xsub}                  & \multicolumn{1}{l}{Xview}                        \\ \hline
3S-CrossCLR &                        & 51.1                         & 50                           \\
3S-AimCLR                           &                        & 54.8                         & 54.3                         \\
3S-SkeleMixCLR                     &                        & 55.3                         & 55.7                         \\
\textbf{Ours}                & \multirow{-4}{*}{1\%}  & {\color[HTML]{FF0000} 59.6} & {\color[HTML]{FF0000} 59.2} \\ \hline
3S-CrossCLR                         &                        & 74.4                         & 77.8                         \\
3S-AimCLR                           &                        & 78.2                         & 81.6                         \\
3S-SkeleMixCLR                      &                        & 79.9                         & 83.6                         \\
\textbf{Ours}                 & \multirow{-4}{*}{10\%} & {\color[HTML]{FF0000} 81.5} & {\color[HTML]{FF0000} 83.8} \\ \hline
\end{tabular}
}
\caption{Semi-supervised evaluation results.}
\label{table5}
\end{table}

\textbf{Finetune evaluation results.}
From our experimental results in Table \ref{table6}, our method has surpassed the three-stream results of some recent methods in only joint-stream, which shows the effectiveness of our method. From the overall effect, our method provides the most effective pre-training parameters for supervised fine-tuning.
\begin{table}[htp]
\centering
\small
{ 
\begin{tabular}{c|c|cc|cc}
\hline
     &   & \multicolumn{2}{c|}{NTU-60} & \multicolumn{2}{c}{NTU-120}    \\  \cline{3-6}
 \multirow{-2}{*}{Method}        & 
  \multirow{-2}{*}{Stream}        &\multicolumn{1}{l}{Xsub}                  & \multicolumn{1}{l|}{Xview}                 & \multicolumn{1}{l}{Xsub}                &  \multicolumn{1}{l}{Xset}                 \\ \hline
SkeletonCLR      &                                 & 82.2                        & 88.9                        & 73.6                        & 75.3                        \\
AimCLR            &                                 & 83.0                        & 89.2                        & 77.2                        & 76.0                        \\
SkeleMixCLR        &                                 & 84.5                        & 91.1                        & 75.1                        & 76.0                        \\
\textbf{Ours} & \multirow{-4}{*}{J} & {\color[HTML]{FF0000} 87.3} & {\color[HTML]{FF0000} 92.8} & {\color[HTML]{FF0000} 77.3} & {\color[HTML]{FF0000} 87.8} \\ \hline
ST-GCN           &                                 & 85.2                        & 91.4                        & 77.2                        & 77.1                        \\
SkeletonCLR     &                                 & 86.2                        & 92.5                        & 80.5                        & 80.4                        \\
AimCLR           &                                 & 86.9                        & 92.8                        & 80.1                        & 80.9                        \\
HiCLR            &                                 & 88.3                        & 93.2                        & 82.1                        & 83.7                        \\
SkeleMixCLR       &                                 & 87.8                        & 93.9                        & 81.6                        & 81.2                        \\
\textbf{Ours} & \multirow{-6}{*}{3S}  & {\color[HTML]{FF0000} 89.4} & {\color[HTML]{FF0000} 94.5} & {\color[HTML]{FF0000} 83.4} & {\color[HTML]{FF0000} 92.7} \\ \hline
\end{tabular}
}
\caption{Fully finetune evaluation results, J means joint, 3S indicates three streams.}
\label{table6}
\end{table}

\subsection{Ablation Study}
Ablation studies are conducted on NTU-60 dataset, and the related evaluation protocol is introduced in Section \ref{subsec: Experimental settings}.

\label{subsec:Ablation}
\begin{table}[htp]
\centering
\small
{ 
    \begin{tabular}{c|c|c|cc}
    \hline
        Negative pair:  $q_{s}$ vs $q_{ns}$ & $L_{s}$ & $L_{ns}$ & Xsub & Xview  \\ \hline
        × & $\surd$ & $\surd$  & 80.9 & 84.1  \\ \hline
        × & $\surd$  & × & 75.5 & 79.2  \\ \hline
        $\surd$  & $\surd$  & × & 76.3 & 80.4  \\ \hline
        $\surd$  & $\surd$  & $\surd$  &  {\color[HTML]{FF0000} 80.3}& 
        {\color[HTML]{FF0000} 86.1}\\ \hline
    \end{tabular}
}
\caption{Ablation study of loss function designs on NTU-60 dataset Joint level.}
\label{table17}
\end{table}
\textbf{Ablation study of local contrastive loss function designs}
In Section \ref{subsec: Local Contrastive Learning}, we introduce $L_{local}$ to optimize local contrastive learning, which is composed of two mirrored loss functions for the salient area and the non-salient area after weighting. In the experiment, we verified the effect of adding $q_{s}$ and $q_{ns}$ as a hard-negative contrastive pair, and the necessity of mirror loss function design for the salient and the non-salient feature area.
Then, the results are shown in Table \ref{table17}. 
In addition, we also conduct ablation experiments for the parameter $\mu$, and verify that the optimal weight $\mu$ of $L_{s}$ and $L_{ns}$ is 0.5, which also shows the mirror image relationship of $L_{s}$ and $L_{ns}$. The experimental results are shown in the \ref{sec: supplementary experiments} together with the parameter tuning experiments of the MHSAM module in Section \ref{subsec: Local Contrastive Learning}.

\section{Conclusion}
\label{sec:conclusion}

In this work, we propose SkeAttnCLR, a novel attention-based contrastive learning framework for self-supervised 3D skeleton action representation learning aimed at enhancing the acquisition of local features. The proposed method emphasizes the importance of learning local action features by leveraging attention-based instance discrimination to bring samples with similar local features closer to the feature space. Experimental results demonstrate that SkeAttnCLR achieves significant improvements over the baseline approach that only relies on global learning. Particularly, our framework achieves outstanding results in various evaluation metrics based on the NTU-60 and NTU-120 datasets.
\bibliographystyle{named}
\bibliography{ijcai23}


\clearpage

\appendix
\section*{Appendix}
In this supplementary material, we provide the following items for a better understanding of the paper:
\begin{itemize}[leftmargin=*]
\item Detailed settings of SkeAttnCLR.
\item More experimental results.
\item t-SNE visualization.
\end{itemize}

\maketitle
\thispagestyle{empty}

\section{Detailed Settings}
\label{sec: detailed architecture}
\textbf{Basic settings.}
Table \ref{table10} is the basic parameters of the contrastive learning framework we use. Table \ref{table11} shows the specific parameters of the SGD optimizer we adopt except those mentioned in the main text of the paper. 
Table \ref{table12} is the parameter configuration of normal data augmentation, which follows the configuration in previous works.
Table \ref{table13} shows the configuration of data mixing augmentation applied in Section 4.2 as extra data augmentation. 
These configurations represent the cropped Skeleton fragments that must be manipulated when data mixing.

\textbf{Backbones settings.}
This section shows the specific configurations of backbones we follow in our experiments. Table \ref{table14} is the configuration of ST-GCN mainly used in our experiments. In order to ensure a fair comparison of experimental results, we conduct experiments using the same configuration of ST-GCN as the backbone. Table \ref{table15} and Table \ref{table16} are the specific parameters of BIGRU and Transformer (DSTA) used in the main text compared with the baseline, which aims to explore the generalizability of the proposed methods with different types of encoders.

\begin{table}[htp]
\centering
\small
{ 
\begin{tabular}{c|cl}
config      & \multicolumn{2}{c}{value} \\ \hline
feature dim & \multicolumn{2}{c}{128}   \\
queue size  & \multicolumn{2}{c}{32768} \\
momentum(M) & \multicolumn{2}{c}{0.996} \\
temperature & \multicolumn{2}{c}{0.2}   \\
lambda($\lambda$)   & \multicolumn{2}{c}{2}     \\
mu($\mu$)       & \multicolumn{2}{c}{0.5}  
\end{tabular}
}
\caption{Parameters setting for contrastive learning framework.}
\label{table10}
\end{table}

\begin{table}[htp]
\small
\centering
{ 
\begin{tabular}{c|cl}
config                           & \multicolumn{2}{c}{value}                            \\ \hline
nesterov & \multicolumn{2}{c}{FALSE}                            \\
weight\_decay                    & \multicolumn{2}{c}{1.00E-04} \\
base\_lr & \multicolumn{2}{c}{0.1}                             
\end{tabular}
}
\caption{SGD optimizer settings.}
\label{table11}
\end{table}


\begin{table}[htp]
\centering
\small
{ 
\begin{tabular}{c|cl}
config                       & \multicolumn{2}{c}{value} \\ \hline
window size                  & \multicolumn{2}{c}{64}    \\
shear amplitude              & \multicolumn{2}{c}{0.5}   \\
temperal\_padding\_ratio     & \multicolumn{2}{c}{6}     \\
mmap & \multicolumn{2}{c}{TRUE} 
\end{tabular}
}
\caption{Norm data augmentation settings.}
\label{table12}
\end{table}

\begin{table}
\centering
\small
{ 
\begin{tabular}{c|cl}
config                                & \multicolumn{2}{c}{value}      \\ \hline
spacial\_l                            & \multicolumn{2}{c}{3}          \\
spacial\_u                            & \multicolumn{2}{c}{4}          \\
temporal\_l                           & \multicolumn{2}{c}{4}          \\
temporal\_u                           & \multicolumn{2}{c}{7}          \\
swap mode                             & \multicolumn{2}{c}{"swap"}     \\
spatial\_mode & \multicolumn{2}{c}{"semantic"}
\end{tabular}
}
\caption{Data mixing augmentation settings.}
\label{table13}
\end{table}

\begin{table}[htp]
\centering
\small
{ 
\begin{tabular}{c|cl}
config                                              & \multicolumn{2}{c}{value}                                    \\ \hline
in channel                                          & \multicolumn{2}{c}{3}                                        \\
hidden channel                                      & \multicolumn{2}{c}{64}                                       \\
hidden dim                                          & \multicolumn{2}{c}{256}                                      \\
droup out                                           & \multicolumn{2}{c}{0.5}                                      \\
class number                                        & \multicolumn{2}{c}{60 (or 120)}                               \\
edge\_importance\_weighting & \multicolumn{2}{c}{TRUE}                        \\
graph\_args                                         & \multicolumn{2}{c}{Layout: "ntu-rgb+d",
strategy: "spatial"}
\end{tabular}
}
\caption{Parameters setting for ST-GCN.}
\label{table14}
\end{table}

\begin{table}[htp]
\centering
\small
  \resizebox{0.45\linewidth}{!}
{ 
\begin{tabular}{c|cl}
config                                & \multicolumn{2}{c}{value}      \\ \hline
input size                            & \multicolumn{2}{c}{150}        \\
hidden size   & \multicolumn{2}{c}{1024}       \\
layers number & \multicolumn{2}{c}{3}          \\
class number                          & \multicolumn{2}{c}{60 (or 120)}
\end{tabular}
}
\caption{Parameters setting for BIGRU.}
\label{table15}
\end{table}

\begin{table}[htp]
\centering
\small
{ 
\begin{tabular}{c|cl}
config          & value      &  \\ \cline{1-2}
frame number    & 64         &  \\
joint number    & 25         &  \\
input channel   & 3          &  \\
hidden dim      & 256        &  \\
number of heads & 8          &  \\
dropout        & 0.5        &  \\
class number    & 60 (or 120) &  \\
config list &
  \begin{tabular}[c]{@{}c@{}}{[}[64, 64, 16, 1],[64, 128, 32, 2], \\ {[128, 128, 32, 1]},{[128, 256, 64, 2]},\\ {[256, 256, 64, 1]}{]}\end{tabular} &
  
\end{tabular}
}
\caption{Parameters setting for Transformer (DSTA).}
\label{table16}
\end{table}

\section{Supplementary experiment}
\label{sec: supplementary experiments}The supplementary experiments shown in this section all use ST-GCN as the backbone for experiments.
\begin{table}[htp]

\centering
\small
{ 
\begin{tabular}{c|cc|cc}
\hline
 & \multicolumn{2}{c|}{Linear} & \multicolumn{2}{c}{KNN}    \\  \cline{2-5}
 \multirow{-2}{*}{$\lambda$}              &\multicolumn{1}{l}{Xsub}                  & \multicolumn{1}{l|}{Xview}                 & \multicolumn{1}{l}{Xsub}                &  \multicolumn{1}{l}{Xview}                 \\ \hline
0.5        & 80.9                                 & 85.4                                 & 69.1                                & 75.6                                 \\
1          & 80.8                                 & 85.5                                 & 68.2                                 & 76.3                                 \\
\textbf{2} & {\textbf{80.3}} & {\textbf{86.1}} & {\textbf{69.4}} & {\textbf{76.8}} \\
4          & 74.3                                 & 85.6                                 & 59.0                                 & 75.2                                 \\
8          & 80.6                                  & 84.4                                 & 70.3                                 & 73.4                                 \\ \hline
\end{tabular}
}
\caption{Hyperparameter $\lambda$ experiments on NTU-60 dataset.}
\label{table7}
\end{table}
\textbf{Hyperparameter $\lambda$ experiments for soft mask tolerance.}
In manuscript Section \ref{subsec: Local Contrastive Learning}, we have introduced that the hyperparameter $\lambda$ affects the tolerance of the soft mask generated by the MHSAM module to neutral features. In order to verify the influence of different values of $\lambda$ on the training effect, we experiment with five values of 0.5, 1, 2, 4, and 8 respectively in Table \ref{table7}. By comparing the linear evaluation results of joint-level on the NTU-60 dataset, we finally use $\lambda$ with a value of 2 as our regular setting. It demonstrates from the experiment that the soft mask tending to be polarized is not conducive to learning better local features. Besides, if the value of lambda is too small, it also affects the performance negatively.

\begin{table}[htp]
\centering
\small
{ 
\begin{tabular}{c|cc|cc}
\hline
 & \multicolumn{2}{c|}{Linear} & \multicolumn{2}{c}{KNN}    \\  \cline{2-5}
 \multirow{-2}{*}{$\mu$}              &\multicolumn{1}{l}{Xsub}                  & \multicolumn{1}{l|}{Xview}                 & \multicolumn{1}{l}{Xsub}                &  \multicolumn{1}{l}{Xview}                                            \\ \hline
0.1          & 76.4                                 & 80.7                                  & 62.6                                 & 68.0                                \\
0.3          & 77.8                                 & 80.5                                 & 65.8                                 & 67.2                                 \\
\textbf{0.5} & {\textbf{80.3}} & {\textbf{86.1}} & {\textbf{69.4}} & {\textbf{76.8}} \\
0.7          & 78.0                                 & 83.1                                 & 66.0                                 & 72.2                                 \\
0.9          & 76.9                                & 82.6                                  & 62.7                                 & 69.7                                \\ \hline
\end{tabular}
}
\caption{Ablation experiments with parameters $\mu$ on NTU-60 dataset.}
\label{table8}
\end{table}


\textbf{Ablation experiments with local contrastive learning parameters $\mu$.}
Since it is not clear whether the salient area of attention is symmetrical to the non-salient area, the parameter $\mu$ is set to balance the proportion of $L_{s}$ and $L_{s}$ in $L_{\text {local }}$. We experiment with five values of 0.1, 0.3, 0.5, 0.7, and 0.9. The results are shown in Table \ref{table8} and we select the best effect as our regular setting. The experimental results show that the $L_{s}$ and $L_{ns}$ of the feature-level weighting calculation using the mask generated by MHSAM have symmetry.

\begin{table}[htp]

\centering
\small
{ 
\begin{tabular}{c|cc|cc}
\hline
 & \multicolumn{2}{c|}{Linear} & \multicolumn{2}{c}{KNN}    \\  \cline{2-5}
 \multirow{-2}{*}{Heads}              &\multicolumn{1}{l}{Xsub}                  & \multicolumn{1}{l|}{Xview}                 & \multicolumn{1}{l}{Xsub}                &  \multicolumn{1}{l}{Xview}                                           \\ \hline
4                                 & 81.3                                 & 84.9                                 & 69.6                                 & 74.6                                 \\
{ \textbf{8}} & {\textbf{80.3}} & {\textbf{86.1}} & {\textbf{69.4}} & {\textbf{76.8}} \\
16                                & {80.1}          & {83.6}           & {67.6}          & {70.4}          \\
32                                & 80.8                                 & 84.3                                 & 68.0                                 & 72.8                                 \\ \hline
\end{tabular}
}
\caption{Abliation study for multi-head self-attention on NTU-60 dataset.}
\label{table9}
\end{table}

\textbf{Experiments for the number of heads in MHSAM.}
The number of heads is very important for the multi-head self-attention mechanism(MHSAM). We test with 4, 8, 16, and 32 heads respectively in the experiment. As shown in Table \ref{table9}, we select 8 heads with the best effect as our regular setting.



\textbf{Experiment result of transfer learning.}
In order to prove that the proposed method has good transfer learning ability, the experimental results of transfer learning are supplemented and shown in Table \ref{table110}. The proposed method supplements the linear evaluation and finetunes evaluation results of cross-domain transfer learning from NTU60 to PKU-MMD in the experiment where ST-GCN is the backbone. However, due to the lack of relevant experimental results of transfer learning under the same conditions in the previous work, it has not been compared with other methods for the time being. This paper supplements this experiment so that future work can use it as a benchmark for comparison.
\begin{table}[htp]
\centering
\small
{ 
    \begin{tabular}{c|c|c|c|c|c|c|c}
    \hline
        \multicolumn{4}{c|}{Linear evaluation} &  \multicolumn{4}{c}{Finetune evaluation}   \\ \hline
        \multicolumn{2}{c|}{PKU part1} & \multicolumn{2}{c|}{PKU part2} & \multicolumn{2}{c|}{PKU part1} & \multicolumn{2}{c}{PKU part2} \\ \hline
        xsub & xview & xsub & xview & xsub & xview & xsub & xview  \\ \hline
        87.0 & 91.8 & 58.0 & 50.9 & 94.5 & 97.0 & 65.8 & 58.8 \\ \hline
    \end{tabular}
}
\caption{Experiment result of transfer learning by NTU60 to PKU on Joint level.}
\label{table110}
\end{table}
\section{t-SNE Visualization}
\label{sec: visualization}
\begin{figure*}[htp]
  \centering
  \captionsetup{font=small}
  \includegraphics[width=1.0\linewidth]{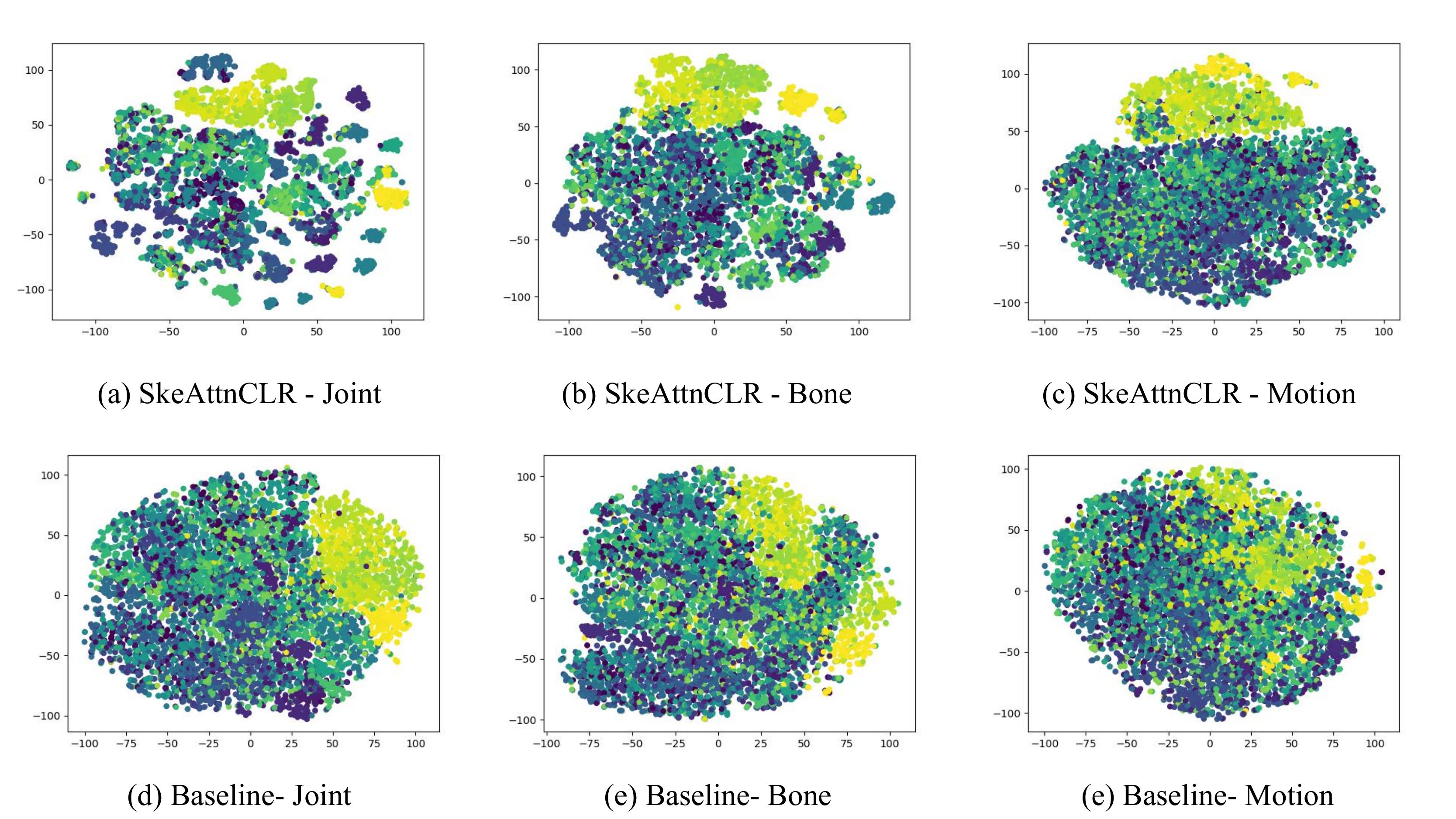}
   \caption { The t-SNE visualization results using the NTU-60 dataset with xsub settings. (a), (b), and (c) indicate the visualizations of SkeAttnCLR in joint, bone, and motion separately. (d), (e), and (f) show the visualization results of the baseline (SkeletonCLR).}
   \label{fig5}
\end{figure*}

\begin{figure*}[htp]
  \centering
  \captionsetup{font=small}
  \includegraphics[width=1.0\linewidth]{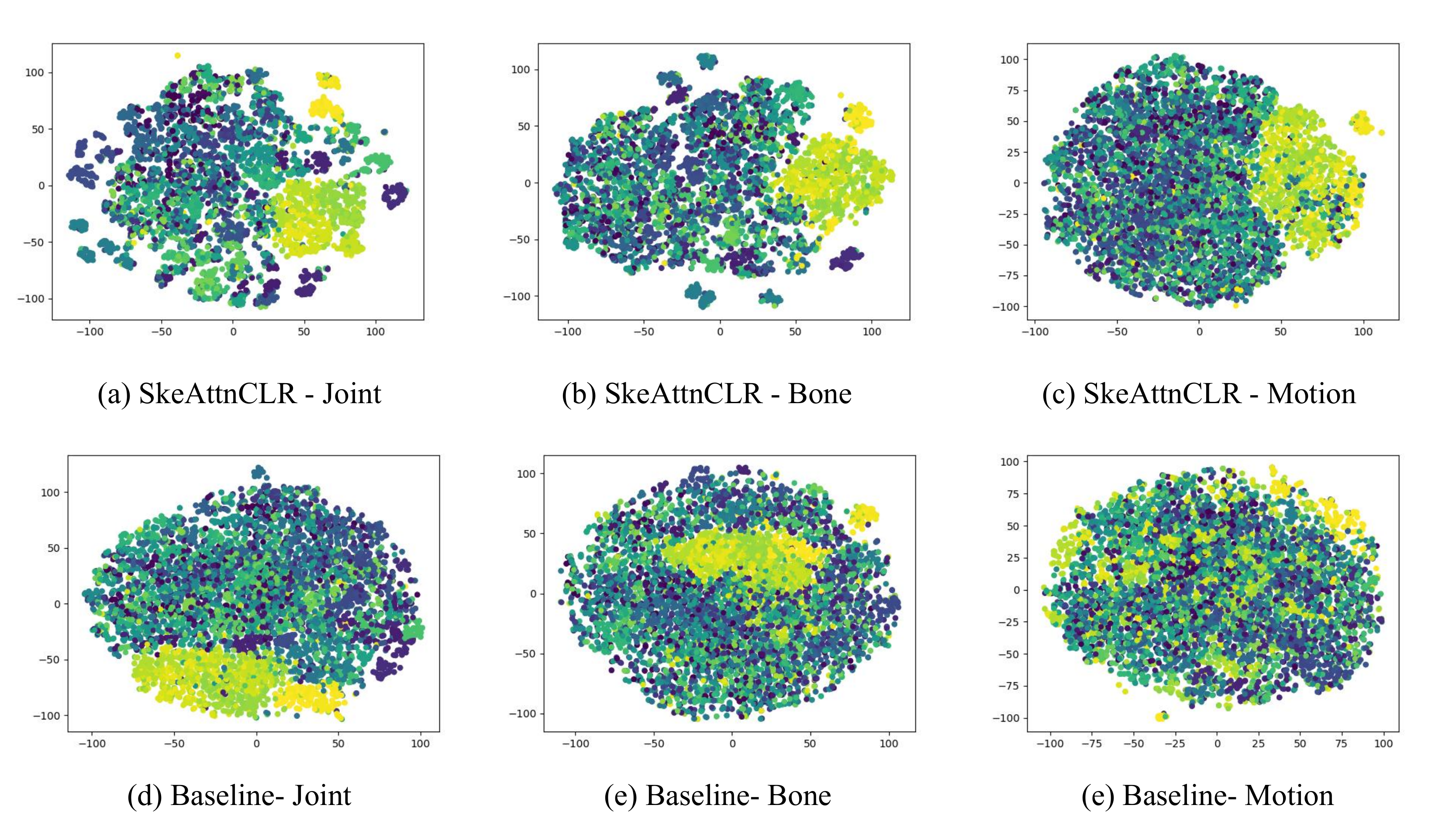}
   \caption { The t-SNE visualization results using the NTU-60 dataset with xview settings. (a), (b), and (c) indicate the visualizations of SkeAttnCLR in joint, bone, and motion separately. (d), (e), and (f) show the visualization results of the baseline (SkeletonCLR).}
   \label{fig6}
\end{figure*}

We provide t-Distributed Stochastic Neighbor Embedding (t-SNE) visualization results of SkeAttnCLR and make comparisons with the baseline (SkeletonCLR) in Fig. \ref{fig5} (with NTU-60 xsub setting) and Fig. \ref{fig6} (with NTU-60 xview setting).  In order to have a clear visualization, we only take 20 $\%$ data from the NTU-60 dataset for embedding estimation. As it showed from the visualizations, SkeAttnCLR achieves better clustering results than the baseline in joint, bone, and motion. The embeddings are clustered closer than that of SkeletonCLR in different settings, which indicates our method has the ability to extract more discriminative features.

\end{document}